\documentclass{edm_article}

\usepackage{graphicx}
\usepackage{siunitx}
\usepackage{pbox}
\usepackage{booktabs}
\usepackage{array}
\usepackage{tabularx}
\usepackage{multirow}
\usepackage{rotating, booktabs}
\usepackage{lscape}
\usepackage{csquotes}
\usepackage{amsfonts}
\usepackage{amsmath}
\usepackage{xcolor}
\usepackage{hyperref}
\usepackage{url}

\begin{document}

\title{Process-BERT: A Framework for Representation Learning on Educational Process Data}

\numberofauthors{1}
\author{
Alexander Scarlatos$^1$, Christopher Brinton$^2$, and Andrew Lan$^1$\\
       \affaddr{University of Massachusetts Amherst$^1$, Purdue University$^2$}\\
       \email{ajscarlatos@cs.umass.edu,cgb@purdue.edu,andrewlan@cs.umass.edu}
}

\maketitle

\begin{abstract}

Educational process data, i.e., logs of detailed student activities in computerized or online learning platforms, has the potential to offer deep insights into how students learn. One can use process data for many downstream tasks such as learning outcome prediction and automatically delivering personalized intervention. However, analyzing process data is challenging since the specific format of process data varies a lot depending on different learning/testing scenarios. In this paper, we propose a framework for learning representations of educational process data that is applicable across many different learning scenarios. Our framework consists of a pre-training step that uses BERT-type objectives to learn representations from sequential process data and a fine-tuning step that further adjusts these representations on downstream prediction tasks. We apply our framework to the 2019 nation's report card data mining competition dataset that consists of student problem-solving process data and detail the specific models we use in this scenario. We conduct both quantitative and qualitative experiments to show that our framework results in process data representations that are both predictive and informative.\footnote{The code for our implementation and experiments can be found here:  \url{https://github.com/alexscarlatos/clickstream-assessments}.}

\end{abstract}


\section{Introduction}

\emph{Student modeling} \cite{vanlehn1988student} is a key research area in educational data mining (EDM) since it produces estimates of individual factors that affect learning outcomes, including knowledge factors and psychosocial factors such as affect and interest. These estimates can be used to inform \emph{personalization}, either by i) providing feedback to help teachers and instructors monitor student progress and intervene if necessary or ii) providing personalized learning activity recommendations directly through digital learning platforms. 
There exist a wide range of student models, from those that analyze student \emph{responses} to questions, such as item response theory \cite{van2013handbook} and models for knowledge tracing \cite{corbett1994knowledge}, to those that analyze student \emph{activity} within digital learning platforms \cite{botelho2017improving,yang2013turn,yao2021stimuli}.

\sloppy
Educational process data, i.e., data that logs detailed student activity in digitized learning/testing environments, offers us an opportunity to look deeper into the process of learning for each individual student. One can use this process data in many ways: First, standalone process data, especially data from intelligent tutoring systems, learning management systems, or massive open online courses (MOOCs), can help us capture student behavioral patterns and predict future learning outcomes \cite{pardos2014affective} or help prevent early dropout \cite{halawa2014dropout}. Second, process data during assessments, such as the dataset used in the 2019 nation's report card (NAEP) data mining competition \cite{naepchal}, can help us reconstruct the exact process behind how students construct their response to a question. This reconstructed process can potentially help us improve our estimate of student knowledge levels more than using only observed response \cite{ckt}. Therefore, due to these potential benefits and the  availability of such data as digital learning becomes prevalent, there is a significant recent interest among the EDM community in analyzing process data, demonstrated by the successful workshop on process analysis methods for educational data at EDM 2021 \cite{processws}. 

\sloppy
One key challenge in educational process data analysis is how to represent process data. This problem is both challenging and important: it is challenging since process data has many different forms, including student activity logs \cite{sherry}, video-watching clickstreams \cite{weiyu}, keystrokes \cite{danielle}, and problem-solving clickstreams \cite{naepchal}. Therefore, existing methods for representing process data are largely domain-specific, i.e., they are developed for a particular form of process data. The problem of how to represent process data is also important since these good representations lead to better performance in downstream tasks, as evident in recent advances in other fields: pre-trained language models such as BERT \cite{devlin2018bert} in natural language processing and contrastive learning-based representations \cite{contrastive} in computer vision. The idea is simple: since we often lack a large amount of labels on the variable of interest in the prediction task, e.g., student learning outcomes, using these labels to learn representations of student process data in a supervised learning setup is not desirable. Instead, we start with using the rich process data itself in a \emph{pre-training} step to learn representations through self-supervised learning before \emph{fine-tuning} these representations in the actual downstream prediction task. This setup has been demonstrated to learn informative representations of raw data, resulting in state-of-the-art downstream prediction accuracy. 

\newpage
\subsection{Contributions}

In this paper, we propose a generic framework for representation learning from educational process data and apply it to the NAEP Competition dataset \cite{naepchal}. Our contributions can be summarized as follows:
\begin{itemize}
    \item We outline a two-stage process for representation learning on process data: First, we detail how to learn process data representations in a pre-training setup using objectives similar to that used in BERT. Our method is applicable to process data that comes in the format of \emph{time series}. Second, we detail how to fine-tune these representations and use them in a downstream supervised learning task, e.g., predicting learning outcomes. 
    \item We apply our framework to a specific type of process data: problem-solving clickstreams as students take an online NAEP assessment.\footnote{Examples of NAEP assessments can be found using the NAEP question tool at \url{https://nces.ed.gov/NationsReportCard/nqt/Search}.} We detail the specific components of the representation learning and downstream prediction models that we use for this dataset. 
    \item We conduct a series of quantitative experiments to show that our framework outperforms existing methods in a series of downstream prediction tasks. We also conduct a series of qualitative experiments to show that our framework is able to learn meaningful process data representations. 
\end{itemize}

\section{Related Work}

There are many existing methods for analyzing educational process data. First, we can use association rule mining \cite{asm} to discover frequent if-then rules among various variables of interest among student activity \cite{merceron2004mining}. This method leads to highly interpretable results but is limited to only binary-valued variables, e.g., whether a student spends more than average time on a learning resource. Second, we can use graphs (see the EDM 2017 Workshop \cite{gedm}) or Markov models \cite{faucon2016semi} to reflect the sequential nature of student activities and extract frequent transition patterns between activity types. These methods also give us an intuitive understanding of student learning processes but are limited to only analyzing the type of activities (e.g., each type is represented by a node in the graph). 

For the prediction of variables of interest from process data, often student learning outcomes, there are two types of existing methods. The first type uses feature engineering \cite{botelho2017improving,kaser,lan2017behavior,pardos2014affective,nirmal,zehner} plus classification, i.e., developing a set of hand-crafted features that summarize student activities as representations of raw process data followed by training a classifier on these features to predict student outcomes. The latter two methods were specifically designed for the NAEP Competition data (see \cite{alina} for background on NAEP process data) and have achieved good results; in fact, the top entries to the competition all rely on feature engineering-based process data representations. The second type uses deep learning to learn latent vectorized representations of raw process data, often using recurrent neural networks \cite{ckt}. These methods excel at predictive accuracy on the downstream prediction task but often lack interpretability to some degree compared to hand-crafted features. Our pre-training setup has some similarities with that proposed in \cite{ckt}, with the main difference being that we use BERT-type objectives while \cite{ckt} uses a sequence-to-sequence autoencoder setup.

\section{Methodology}

In this section, we detail our framework for pre-training and transfer learning on educational process data that encompasses many real-world educational scenarios. The basic ideas behind our framework follow from those in natural language processing (NLP) research but are adapted for student learning process data. There are three main technical components in our framework:
\begin{itemize}
    \item The \textbf{process model}, which takes a student's process data as input and produces its \textit{latent representation vectors} as output. 
    \item The \textbf{pre-training objectives}, which are a series of prediction tasks that we use in a self-supervised pre-training phase on the process data to learn its representations via the process model. 
    \item The \textbf{transfer function}, which adapts the output of the process model, i.e., the latent representations of process data, for use in a downstream learning outcome prediction task.
\end{itemize}

Before we detail the three components individually, we emphasize that our framework can be applied to any educational process data that is in the form of \emph{time series}. Specifically, let $e^i_1, \dots, e^i_{T^i}$ denote the sequence of \emph{events} that reflect student~$i$'s learning process, where each entry $e^i_t = (a^i_t, m^i_t)$ represents the event at a discrete time step~$t$, with a total of $T^i$ time steps. $a^i_t$ and $m^i_t$ are the \emph{type} and (actual) \emph{timestamp} of the event for the student at the time step, respectively. We note that these two aspects are universal to almost any process data format; however, depending on the specific educational scenario, there may also be additional information associated with each event (or groups of events): for problem solving process data, each event can also contain information on the correctness of the student's current attempt \cite{ckt}, while for video-watching process data, each event can also contain information on the relative timestamp within a video \cite{lan2017behavior}. We refer to these additional features as $\mathbf{f}^i_t$. In what follows, for simplicity of exposition, we will discuss the process data for a single student and drop the student index $i$ unless necessary.

\paragraph{Process model} We can represent the process model as a function $P_\theta$ with $\theta$ being its set of parameters, which takes as input the raw event sequence and produces a series of latent event representation vectors as output:
\begin{align*}
    \mathbf{z}_1, \dots, \mathbf{z}_T = P_\theta(e_1, \dots, e_{T}),
\end{align*}
where vector $\mathbf{z}_t$ is the latent representation of $e_t$. We can use many different types of models as the process model, depending on the specific process data; one possible choice is to use recurrent neural networks where $\mathbf{z}_t$ depends on all previous events, $e_1, \ldots, e_{t}$. Another possible choice is to use attention-based models like the Transformer \cite{tf} that can provide fully contextualized representations where $\mathbf{z}_t$ depends on all previous and future events, $e_1, \ldots, e_{T}$. Once we pre-train the model using a series of objectives on the process data itself, we refer to the parameters as $\theta^*$.

\paragraph{Pre-training objectives} Since the amount of data labels accompanying process data, e.g., learning outcome indicators, can be limited, learning good process data representations using these labels may be unrealistic. Therefore, we design a series of pre-training objectives defined on the process data only to help us learn a good representation, $P_\theta(\cdot)$, an approach known as \emph{self-supervised learning}. In our framework, we employ learning objectives similar to those in BERT \cite{devlin2018bert}: we predict $a_t$ and $m_t$ using some subset of $\mathbf{z}_1, \dots, \mathbf{z}_T$. For different process model choices, this subset should be chosen differently to prevent information on the current event being leaked into its prediction. 
Moreover, depending on the specific educational scenario, we may include additional objectives for our pre-training task using additional event features $\mathbf{f}_t$.  
Formally, we define our pre-training objective, $\mathcal{L}_{PT}$, as
\begin{align*}
    \mathcal{L}_{PT} = \sum_{t \in \Omega \subseteq \{1,\ldots,T\}} \left( \mathcal{L}_a(a_t, \hat{a}_t) + \mathcal{L}_m(m_t, \hat{m}_t) + \mathcal{L}_{\mathbf{f}}(\mathbf{f}_t, \hat{\mathbf{f}}_t) \right),
\end{align*}
where $\mathcal{L}_a$, $\mathcal{L}_m$, and $\mathcal{L}_{f^j}$ denote the event type, timestamp, and (optional) additional feature prediction losses, respectively, and $\hat{a}_t$, $\hat{m}_t$, and $\hat{\mathbf{f}}_t$ denote predicted values. By minimizing this pre-training objective, we aim at learning good representations of raw process data that are predictive of the events.

\paragraph{Transfer function} After the process model is pre-trained, we aim at transferring the representations it learned to downstream tasks on learning outcome prediction where $y$ is the label for a student's learning outcome. 
We use a transfer function, $Q_\phi$, that takes as input a student's event sequence representations and predicts the learning outcome, $\hat{y}$, as output, i.e.,
\begin{align*}
\hat{y} = Q_\phi(P_{\theta}(e_1, \dots, e_T)).
\end{align*}
We train this transfer function on the downstream prediction task by minimizing the prediction loss on the learning outcome label, $\mathcal{L}_{FT}(y,\hat{y})$. Specifically, we learn the parameters of the transfer function, $\phi$, from scratch, while we fine-tune the parameters of the process model, $\theta$ (initialized to $\theta^*$), by backpropagating the gradients. Alternatively, we may choose to freeze the process model's parameters and not update it while only learning $\phi$ during the fine-tuning phase.

\section{Application to NAEP Process Data}

In this section, we apply our methodology to several learning outcome prediction tasks using the NAEP 2019 competition dataset \cite{naepchal}. The dataset contains clickstream logs from students working on two blocks of online NAEP assessments. Students are given a time limit of 30 minutes per block, in which they can complete questions in any order. The questions vary in type, including multiple choice, matching, fill in the blank, and ``mixed'' types. Each event in the log represents a single student action, such as selecting an option in a multiple choice question, typing a character into an answer field, turning on the calculator tool, using the navigation bar to proceed to the next question, etc. Each raw event in the log contains the student ID, question ID, question type, event type, timestamp, and possibly additional event-specific information such as the target of a click event or the field and character of a key press event.
To apply our framework to the NAEP assessment scenario, we process each event so that it has the form $e_t = (a_t, m_t, q_t, c_t)$, where $a_t$ is the event type, $m_t$ is the number of seconds since the student started the test, $q_t$ is an identifier for the current question, and $c_t$ is a response status, which we describe in more detail next. Note that $\mathbf{f}_t$ now consists of $q_t$ and $c_t$ since these features are relevant to the NAEP assessment scenario of question responding activity. We additionally define a \textit{visit} to a question as a contiguous sequence of events that are part of the same question, which is relevant since students may return to previously visited questions at any time within a block.

We define the response status $c_t$ for each event to be either \textit{correct}, \textit{incorrect}, or \textit{incomplete}. For each event during a student's visit to a question, we maintain a state indicating the current response and update it as the student performs actions. At the last event of the visit, if all fields have been filled out, we assign \textit{correct} or \textit{incorrect} by comparing the response state to acceptable answers. If some relevant fields have not been filled out, we assign \textit{incomplete}. We copy the response status of the final event in a visit and assign it to every event in the visit, rather than adjusting the response status throughout the visit, in order to handle potential ambiguity between incomplete and incorrect responses. We collected correct answers from the online NAEP question tool except for the five questions that are not available there. For these, as suggested by \cite{osti_10190399}, we took the majority answer among the top 100 performing students to be correct, which yielded 78-98\% agreement.

\subsection{Process Model}

We use a Bidirectional Long Short-Term Memory network (Bi-LSTM) \cite{lstm} as our process model. An LSTM is a popular type of RNN that is often used for time series data. A bidirectional LSTM runs two independent LSTM models in parallel: a \textit{forward} LSTM that processes the sequence in order, and a \textit{backward} LSTM that processes the sequence in reverse. The forward LSTM's hidden state at time step $t$, $\overrightarrow{\mathbf{h}}_t$, depends on all previous events, i.e., $e_0,\ldots,e_t$. The backward LSTM's hidden state at time step $t$, $\overleftarrow{\mathbf{h}}_t$, depends on information from all future events, i.e., $e_{t},\ldots,e_T$. Therefore, the output of the process model is this set of contextualized representations of all events, $\{\overrightarrow{\mathbf{h}}_1, \dots, \overrightarrow{\mathbf{h}}_T, \overleftarrow{\mathbf{h}}_1, \dots, \overleftarrow{\mathbf{h}}_T$\}. We also experimented with a Transformer encoder model, but found it to be less effective than the Bi-LSTM on our dataset, possibly due to overfitting on the NAEP competition data with limited scale.

Since the input to the Bi-LSTM model at each time step must be a vector, we introduce a vectorized form, $\mathbf{e}_t$, of each event. We use learnable embeddings $emb_a$ and $emb_q$ to represent $a_t$ and $q_t$ respectively, and one-hot encoding to represent $c_t$. We concatenate these representations to form $\mathbf{e}_t = (emb_a(a_t), emb_q(q_t), onehot(c_t), m_t)$.
We chose to use one-hot encoding for $c_t$ instead of embeddings since there are only three possible correctness states.

\subsection{Pre-training Objectives}

\begin{figure}
    \centering
    \includegraphics[width=2.5in]{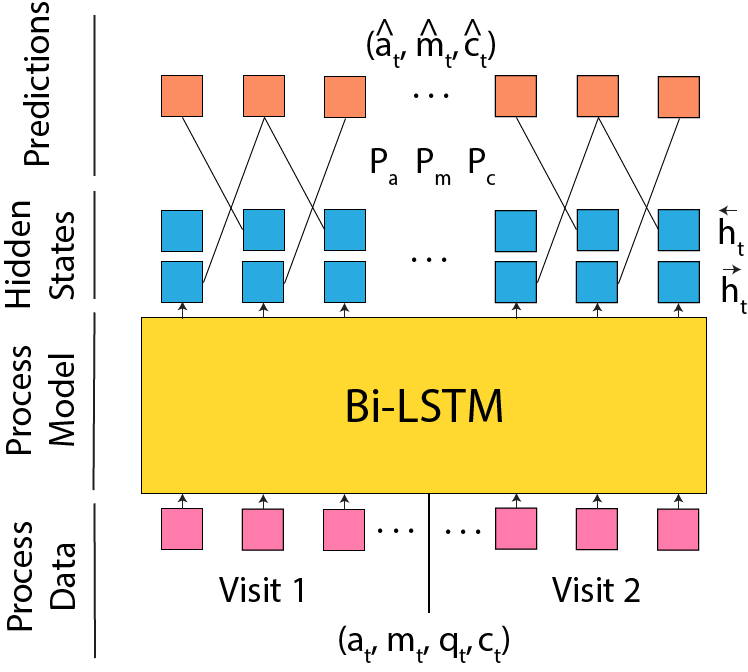}
    \caption{The process model is pre-trained by predicting the properties of each event given surrounding context.}
    \label{fig:pretraining}
\end{figure}

We now detail the objectives that we use to pre-train our process model.
We design separate objectives to predict the \textit{event type}, \textit{timestamp},  and \textit{response status} of each event. We design the process model to accept the events of a single question at a time, but use a single model to represent all questions, rather than a separate model for each question. Note that we don't include a pre-training objective to predict the question ID of an event; this is because $q_t$ is the same for every event in a sequence, so predicting it would not leverage any additional information. The flow of data from input sequence to prediction is shown in Figure \ref{fig:pretraining}; the way these predictions are generated, as well as the loss functions used to train them, will be described throughout the rest of this section.

\paragraph{Event Type Prediction} 
To predict the event type, $a_t$, at each time step, we leverage the full context of $a_t$, which includes every event in the input excluding $\mathbf{e}_t$, i.e., $\{\mathbf{e}_1,\ldots,\mathbf{e}_{t-1},\mathbf{e}_{t+1},\ldots,\mathbf{e}_T\}$. Since $\overrightarrow{\mathbf{h}}_{t-1}$ uses on information from $\{\mathbf{e}_1,\ldots,\mathbf{e}_{t-1}\}$ and $\overleftarrow{\mathbf{h}}_{t+1}$ uses on information from $\{\mathbf{e}_{t+1},\ldots,\mathbf{e}_T\}$, we use $\mathbf{z}_t=(\overrightarrow{\mathbf{h}}_{t-1},\overleftarrow{\mathbf{h}}_{t+1})$, which includes information on all events except $e_t$, to predict $a_t$. 
We pass $\mathbf{z}_t$ through a linear prediction head, $P_a$, use the softmax function \cite{dlbook} to get a probability distribution over possible event types, and use cross-entropy (CE) to calculate the loss. The event type prediction loss is formally given by
\begin{align*}
    \mathcal{L}_a(\mathbf{e}_t) = CE(softmax(P_a(\mathbf{z}_t)), a_t).
\end{align*}

\paragraph{Timestamp Prediction} For the timestamp prediction objective, we use a similar setup as event type prediction. We select the same process model outputs for the predictive state, $\mathbf{z}_t=(\overrightarrow{\mathbf{h}}_{t-1},\overleftarrow{\mathbf{h}}_{t+1})$. We then use a linear prediction head, $P_m$, to obtain a single output value. Then, instead of using $m_t$ directly as the prediction target, we define a new target called the \textit{time ratio}, which we define as
\begin{align*}
    r_t=\frac{m_t-m_{t-1}}{m_{t+1}-m_{t-1}}.
\end{align*}
The time ratio represents at what portion of time $m_t$ occurs between the prior and following events, and will always have a value between 0 and 1. Since $r_t$ is undefined at the first and last time steps of a sequence, we set its values there to 0 and 1, respectively. We then use binary cross-entropy (BCE) to compute the timestamp loss as
\begin{align*}
    \mathcal{L}_m(\mathbf{e}_t) = BCE(\sigma(P_m(\mathbf{z}_t)), r_t),
\end{align*}
where $\sigma$ is the \textit{sigmoid} function. The advantage of this loss over mean squared error (MSE) with exact timestamp as the prediction target is that since time lapses between student events can be very short or very long, the time prediction loss for each event is treated more equally, without having to adjust to the scale of the time lapse.

\paragraph{Response Status Prediction} We predict response status the same way as we predict event type: using a linear prediction head on top of $\mathbf{z}_t$ to predict $c_t$ and minimizing the corresponding CE loss, $L_c(\mathbf{e}_t)$. Unlike event type and timestamp, the model can directly infer response status from the input since it is given as input for any event in a given visit. Nevertheless, we found this prediction objective helpful since it effectively encodes the response status in the process model's output, which benefits downstream prediction tasks we study in this paper.

Combining the event type, timestamp, and response status prediction losses, we arrive at the final pre-training objective for a student's event sequence for an individual question:
\begin{align*}
    \mathcal{L}_{PT} = \sum_{t \in \{1,\ldots,T\}} \mathcal{L}_a(\mathbf{e}_t) + \mathcal{L}_m(\mathbf{e}_t) + \mathcal{L}_c(\mathbf{e}_t).
\end{align*}
By minimizing $\mathcal{L}_{PT}$, we train the process model to reconstruct its input using surrounding context in a self-supervised learning setup, thus encoding relevant information in its latent states as a result. Next, we show how we transfer the learned process model and latent event representations to downstream prediction tasks.

\subsection{Transfer Learning}

\begin{figure*}
    \centering
    \includegraphics[width=5in]{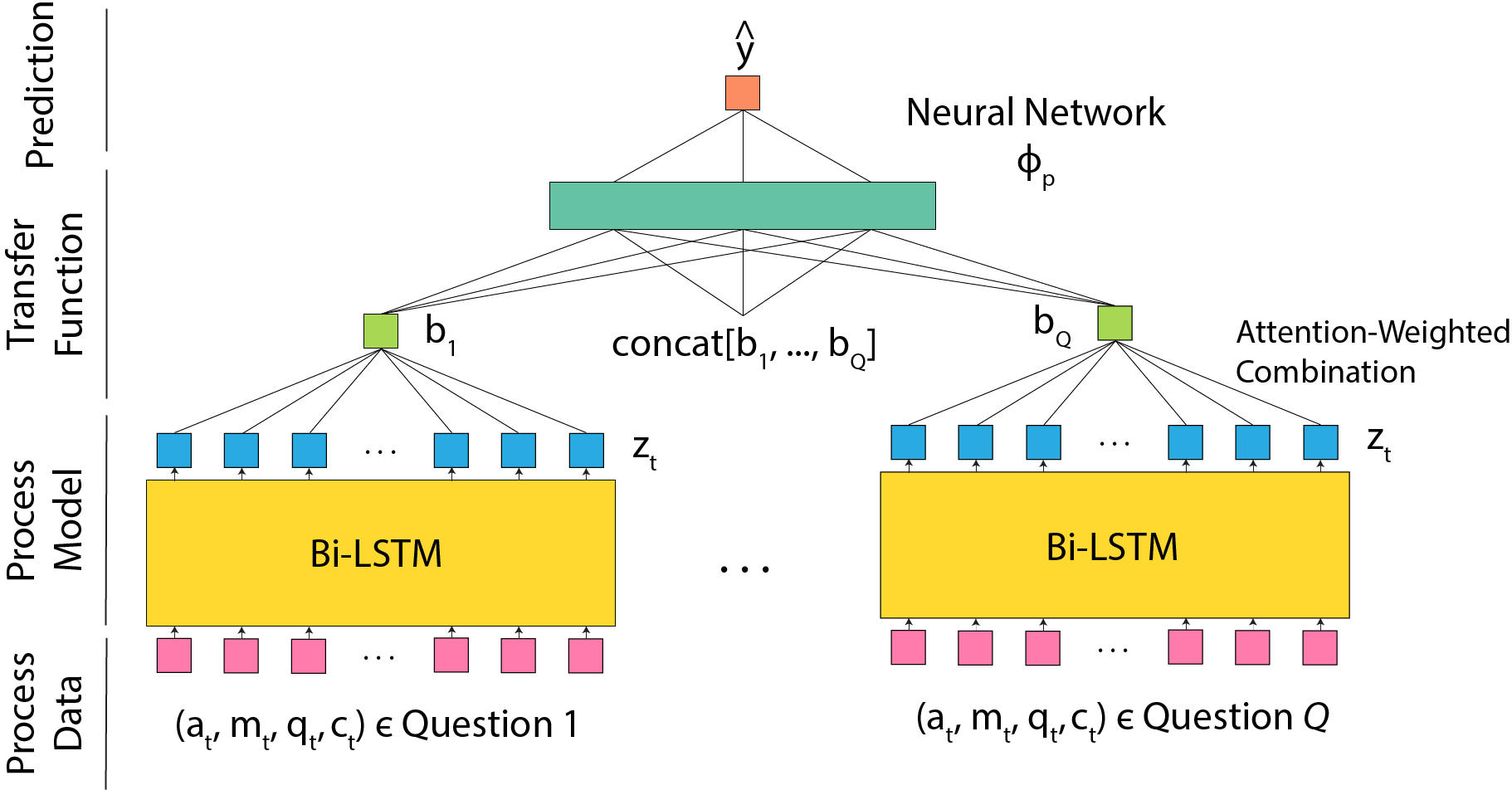}
    \caption{The transfer function, which includes an attention module and neural network, uses the process model's output for each question to generate a final prediction for a student.}
    \label{fig:transfer}
\end{figure*}

We now detail a transfer function, $Q_\phi$, that produces a fixed-size output for a downstream prediction task, given the latent states of the process model as input. For the purpose of transfer learning, we define the outputs of the process model to be $\mathbf{z}_1,\ldots,\mathbf{z}_T=((\overrightarrow{\mathbf{h}}_1, \overleftarrow{\mathbf{h}}_1),\ldots,(\overrightarrow{\mathbf{h}}_T, \overleftarrow{\mathbf{h}}_T))$. This setup results in each $\mathbf{z}_t$ containing contextualized information that is relevant to the input at time step $t$, as was ensured by the pre-training process. The task now becomes how to use each $\mathbf{z}_t$ in a way that will minimize prediction loss for the downstream prediction task. For architectures that use RNNs, it is common to use the final hidden state of the RNN as the input to a linear layer to make predictions, which in our case would be $\mathbf{z}_T$ for the final forward LSTM state and $\mathbf{z}_1$ for the final backward LSTM state. However, RNNs tend to suffer from a bottleneck problem, where hidden states tend to remember information that they obtained recently and forget information from the distant past \cite{attention}. We can circumvent this issue by using all $\mathbf{z}_t$s.
We achieve this by using a learnable attention module \cite{dlbook}, which assigns a weight to each output latent state of the process model, and then uses the weights to combine all outputs into a single vector. We define the weights for the output latent state vectors as
\begin{align*}
    w_1,\ldots,w_T \sim softmax(\phi_w(\mathbf{z}_1),\ldots,\phi_w(\mathbf{z}_T)),
\end{align*}
where $\phi_w$ is a learnable linear projection. We then generate a single vector $\mathbf{b}$ to represent the entire sequence as
\begin{align*}
    \mathbf{b} = \sum_{t \in \{1,\ldots,T\}} w_t \cdot \mathbf{z}_t.
\end{align*}
We can now use this representative vector in a variety of different ways to solve downstream tasks. In the simple case, where we predict the learning outcome on a single question, e.g., the correctness of the student's response to the question, we can use a learnable linear projection $\phi_p$ on $\mathbf{b}$ to predict
\begin{align*}
    \hat{y} = \phi_p(\mathbf{b})
\end{align*}
using events on this question. 

Additionally, in some cases, we may want to predict the learning outcome for each student, which requires us to combine multiple question event sequences for each student; this workflow is shown in Figure \ref{fig:transfer}. For scenarios where each student responds to the same set of questions, which is the case with our dataset, we can use the following approach: 
We concatenate the aggregated question-level latent event representations as $\mathbf{b} = [\mathbf{b}_1, \ldots,  \mathbf{b}_Q]$, where $\mathbf{b}_j$ corresponds to the student's activity on question $j$ and $Q$ is the total number of questions. We use this vector as input to the neural network, $\phi_p$, to predict the learning outcome as
\begin{align*}
    \hat{y} = \phi_p(\mathbf{b}).
\end{align*}
We note that $\hat{y}$ can have any dimension or type: it may be compared directly to the label of a downstream task, or it may be used in the computation of another prediction task, as we detail in the next section. As long as we define a differentiable loss function on $\hat{y}$, we can learn the parameters of $Q_\phi$ and $P_\theta$ from data.

\subsection{Item Response Theory}

A popular framework for making predictions at the question level is item response theory (IRT) \cite{lordirt}; we now detail how to enhance it with behavioral data. The goal of IRT (the 1PL version) is to learn \textit{ability} attributes for students and \textit{difficulty} attributes for questions and use them to make predictions of student performance on questions. Given a series of scored student responses, the attributes can be learned by maximizing the likelihood of
\begin{align*}
    P(Y_{ij} = 1) = \sigma(k_i - d_j),
\end{align*}
where $Y_{ij}$ is 1 when student $i$ gets question $j$ correct and 0 otherwise. $k_i$ is the ability of student $i$ and $d_j$ is the difficulty of question $j$. We learn these attributes by minimizing the binary cross-entropy loss between the predicted probability and the actual correctness label.

We propose a way to augment IRT by adding a new \textit{behavior} term, which adjusts the predicted correctness probability based on observed behavioral data from the student.
Specifically, the formulation becomes
\begin{align*}
    P(Y_{ij} = 1) = \sigma(k_i - d_j + B_{ij})
\end{align*}
where $B_{ij}$ is a scalar representation of the behavior of student $i$ while solving question $j$. We set $B_{ij}$ to be the target of a transfer learning task, and train a predictor for it using our transfer learning approach as
\begin{align*}
    B_{ij} = Q_\phi(P_\theta(\{e^i_t\}_{t: q^i_t = j})).
\end{align*}
By minimizing the binary cross-entropy loss, we train $Q_\phi$ and $P_\theta$ in tandem with $k_i$ and $d_j$. We must also remove response status from $e^i_t$ and not perform the response status pre-training objective because the response status contains correctness information, which is the goal of this task.

It is important to note that we choose IRT models to augment with behavioral data since it is suitable for NAEP assessments; one can safely assume that student knowledge/ability remains constant during the assessment. In other scenarios such as continuous learning while responding to questions, previous work has studied how to augment knowledge tracing models with behavioral data \cite{ckt}. 

\section{Experiments}

In this section, we present the experimental results of our framework applied to the NAEP competition data on several learning outcome prediction tasks. We compare our framework to several existing baselines for both the NAEP process data and other process datasets. We finally investigate the interpretability of the process model's latent representations using visualizations and qualitative analysis.

\subsection{Baselines}

We describe two alternate methods that we compare our framework against: one feature engineering-based method, and another autoencoder-based method.

\paragraph{Feature Engineering}
Feature engineering (FE) is a prediction technique where one manually defines a set of features that summarize a student's process data, e.g., the average amount of time spent per question, followed by training a classifier on top of these features for the prediction task. This technique has found tremendous success on many types of educational data, although a disadvantage is that significant engineering effort needs to be spent on defining and implementing these features, whereas some neural network-based architectures can automatically learn (uninterpretable) features.

To compare FE to our framework, we used the method of the 2nd place submission from the NAEP 2019 competition \cite{nirmal}. Their technique calculates a large number of features for each student activity sequence, most of which are related to timing and correctness. They then use a genetic algorithm (GA) to select the best set of features for a target prediction label. Finally, using the optimal feature subset, they train a large group of models and ensemble them to produce a final prediction for the task. We re-ran their GA and ensemble algorithms to obtain predictions for our per-student prediction label.

\paragraph{Autoencoders} To compare our framework to an existing neural network-based method in a similar setting, we adapted the CKT technique developed in \cite{ckt}, which uses student problem solving process data for the downstream task of knowledge tracing (KT). 
Their technique uses an autoencoder training setup to generate a single latent vector for each student's activity sequence on a single question. Specifically, it channels the final hidden state of an encoder RNN, after all input events have been processed, into a bottleneck vector, which serves as the representative state for the sequence. The bottleneck vector is then fed to the starting state of a decoder RNN that reconstructs the input sequence, and the full model is trained on the reconstruction loss.
They then use these learned representations as features in KT models.
We note that their problem setting is different from NAEP assessments, which is a testing environment: students face a time limit for the entire set of questions and may jump back and forth between questions. 
On the other hand, CKT is designed for student activity spanning over several months, where a student finishes working on a problem before moving onto the next. 
To adapt their method to the NAEP process data, we apply a transfer function that is similar to ours by concatenating the encodings of each question and passing the resulting vector through a fully-connected neural network. 
We also experimented with their approach for downstream tasks that leverages the sequential nature of question visits but found that our transfer function resulted in significantly better performance.

\subsection{Per-Student Labels}

We first evaluate our method on two sets of per-student labels, each of which represents some property of a student's performance in the second block of the exam, predicted using data in the first block only. We refer to the first block as block A and the second as block B. We note that we don't evaluate our method on the NAEP competition label because it yielded high variance in test results and a large difference between validation and test performance in experiments; its usefulness and interpretability is also questionable, as noted by \cite{osti_10190399}. We thus designed the following two labels for evaluation:
\begin{itemize}
    \item The \textbf{score label}, which is a binary indicator of if a student scored above or below the average for block B.
    \item The \textbf{per-question label}, which is a vector of binary indicators of if a student was correct or incorrect for each question in block B.
\end{itemize}

There are 2,464 students in the full dataset, each of which has a sequence for block A and a sequence for block B. In the original competition setup, participants were only provided block A data, where half of the sequences were intended for training and the other half were intended for validation. Additionally, the validation set was split into 3 equal-sized partitions - one where only the first 10 minutes of each sequence are provided, one where only the first 20 minutes are provided, and one where the full 30 minutes are provided. However, we have access to all full student activity sequences on blocks A and B so we use this information to compute the score and per-question labels.

\paragraph{Experimental Setup} For our model and the CKT baseline, we use the following experimental setup. We use the training set provided by the competition as our training and validation sets, and use the 30-minute partition of the validation set provided by the competition as our test set. For each label, we perform 5-fold cross-validation, stratified on the label to ensure the balance between the train and validation sets. 
Our training process contains multiple phases: for each fold and for each phase, we evaluate the model on the validation set at each epoch and use the model with the lowest validation loss. 
We first split each student sequence into sub-sequences, one for each question, where the events of individual visits per question are concatenated. 
We then pre-train the process model on these per-question sequences. 
Next, we freeze the parameters of the process model and train the transfer function on the block B student performance label using the aggregated  representations across all questions for each student. 
After this phase, we fine-tune the process model by un-freezing its parameters and training both the transfer function and the process model on the label. 
We then evaluate the final model on the test set. 

The metric that we use is the area under the receiver operating characteristic curve (AUC), which the standard metric for binary classification that measures how well a model predicts true positives while minimizing false positives. 
The metric is bounded between 0 and 1, with higher values representing better performance. 
For the per-question label, we take the AUC to be the macro average of the AUC for each question in the vector. 
For each label, we collect the AUC on the test set after each fold, and report the average and standard deviation of these AUC numbers.

The FE baseline uses an ensemble where each model is trained separately in 10-fold cross-validations on the training set for three iterations. The ensemble then produces a single prediction for the label on the test set, aggregated over each cross-validation run. Because of this workflow, no validation AUC is available and no standard deviation on the test AUC is available. Additionally, their method only supports a binary label per student, so we only provide results on the score label.

\paragraph{Results and Discussion}

\begin{table}[]
    \centering
    \begin{tabular}{|l|c|c|}
        \hline
        Model & Test AUC & Test AUC without $c_t$ \\
        \hline
        FE & 0.828 & -- \\
        CKT & $0.854 \pm 0.005$ & $\mathbf{0.797} \pm 0.010$ \\
        Ours & $\mathbf{0.868} \pm 0.008$ & $0.792 \pm 0.004$ \\
        \hline
    \end{tabular}
    \caption{We report the AUC of predictions on the score label, as well as the AUC when response status is removed.}
    \label{tab:score}
\end{table}

For the score label, as shown in the second column of Table~\ref{tab:score}, both CKT and our model outperformed the FE baseline. This observation fits our expectation: since sequential neural models have direct access to the raw process data, they are able to pick up on subtleties that may not be captured by human-engineered features. Our method slightly outperforms CKT on this label, indicating that our process model and transfer function are able to capture more information that is indicative of student performance than CKT. We also examine the ability of the models to predict performance strictly using behavioral information, without any indication of correctness. To do this, we repeat the experiments but remove the response status $c_t$ from the input and do not include the associated pre-training objective. We see from the third column of Table~\ref{tab:score} that the AUC drops, as expected, but is still considerably high, indicating that these models can associate student performance with their behavior. With the response status removed, CKT slightly outperforms our model, although the difference is not statistically significant.

\begin{table}[]
    \centering
    \begin{tabular}{|l|c|c|}
        \hline
        Model & Test AUC & Test AUC without $c_t$ \\
        \hline
        CKT & $0.725 \pm 0.005$ & $0.674 \pm 0.006$ \\
        Ours & $\mathbf{0.742} \pm 0.006$ & $\mathbf{0.702} \pm 0.006$ \\
        \hline
    \end{tabular}
    \caption{We report the AUC of predictions on the per-question label, as well as the AUC when response status is removed.}
    \label{tab:per-question}
\end{table}

For the per-question label, as seen in Table~\ref{tab:per-question}, we observe lower AUC numbers than the score label across all models and that AUC decreases when response status is not used. The lower overall performance for this label is as expected since fine-grained labels are generally harder to predict than ones at a higher level. For this label, our model slightly outperforms CKT, both with and without response status.

\subsection{Item Response Theory}

We now evaluate our model in the IRT setting in order to examine if our methodology can improve performance prediction on questions using additional data on student behavior within a question.

\paragraph{Experimental Setup}

For the IRT experiments, we use all student activity sequences across blocks A and B, which we split into per-question sub-sequences. We reserve 20\% of these for a test set; since some students do not visit certain questions, we do a stratified split on student ID and question ID so that the individual ability and difficulty parameters can be sufficiently trained and tested. We perform the split using an iterative multi-label stratification algorithm, which resulted in all questions appearing in both the train and test set, and all but one student appearing in the test set as well as the train set. We perform 5-fold cross-validation on the remaining train set, stratifying the splits again using student ID and question ID. Similar to our setup for the per-student labels, for each fold we pre-train the process model, train the transfer function, fine-tune the model, and then test its performance on the test set. We report the average and standard deviation of the test AUC across these folds. For the base IRT model, since there is no process data involved, we do not perform a pre-training or fine-tuning phase.

\paragraph{Results and Discussion}

\begin{table}[]
    \centering
    \begin{tabular}{|l|c|c|}
        \hline
        Model & All Questions & Completed Questions \\
        \hline
        Base & $0.824 \pm 0.001$ & $0.823 \pm 0.001$ \\
        CKT & $\mathbf{0.836} \pm 0.000$ & $\mathbf{0.830} \pm 0.001$ \\
        Ours & $\mathbf{0.836} \pm 0.002$ & $0.828 \pm 0.001$ \\
        \hline
    \end{tabular}
    \caption{We report the AUC of predictions on the IRT task, as well as the AUC when incomplete questions are not considered.}
    \label{tab:irt}
\end{table}

We see from Table~\ref{tab:irt} that the behavioral data leveraged by both our model and CKT result in a small improvement in test AUC over the base IRT model. We also observe that questions that were left incomplete by students were very easy to predict as incorrect with behavioral data since certain event types missing in a student activity sequence clearly indicate incomplete status. To account for this observation, we also report the AUC after removing incomplete questions from the test set. We see that the performance drops for all models, although more significantly for the behavior-enhanced models, leaving our method's performance slightly below CKT. However, the fact that the behavior-enhanced models still improve over the base IRT model suggests that student behavior provides important additional information on student performance beyond the original student ability and question difficulty parameters in IRT.

\subsection{Ablation Study}

We perform an ablation study of our model on the score label prediction task to examine how different components of our framework affect downstream performance. We experiment with the following modifications, for each of which we repeat the cross-validation experiment on the score label:
\begin{itemize}
    \item Skip the \textbf{event type prediction} pre-training objective.
    \item Skip the \textbf{time prediction} pre-training objective.
    \item Skip the \textbf{response status} pre-training objective.
    \item Skip \textbf{all pre-training objectives} and train the process model and transfer function from scratch on the score label.
    \item Replace the \textbf{attention}-weighted question representations with the concatenation of the final hidden states of the forward and backward LSTMs.
    \item Skip the \textbf{fine-tuning} phase and evaluate the model after training the transfer function on the frozen process model outputs.
\end{itemize}

\begin{table}[]
    \centering
    \begin{tabular}{|l|c|}
        \hline
        Remove & Test AUC \\
        \hline
        None & $0.868 \pm 0.008$ \\
        Event Type Prediction & $0.868 \pm 0.004$ \\
        Time Prediction & $0.867 \pm 0.004$ \\
        Response Status Prediction & $0.820 \pm 0.007$ \\
        All Pre-Training Objectives & $0.819 \pm 0.008$ \\
        Attention & $0.862 \pm 0.005$ \\
        Fine-Tuning & $0.869 \pm 0.008$ \\
        \hline
    \end{tabular}
    \caption{We show how removing pre-training objectives and other model components affects AUC on the score label.}
    \label{tab:ablation-score}
\end{table}

We see from Table~\ref{tab:ablation-score} that score label prediction performance does not suffer from skipping the event type and time prediction pre-training objectives, but does suffer from removing the response status pre-training objective. Removing the attention step for aggregating per-event representations results in a small drop in performance. We also see that fine-tuning does not improve score prediction performance, implying the process model encodes all necessary information in its latent states to make predictions on the score label. 
These results suggest that the primary indicator of student performance on block B is unsurprisingly their performance on block A, indicated by their response correctness; if this information is readily available, the model does not need extra information to make highly accurate predictions on the score label. However, as we saw earlier in Table~\ref{tab:score}, our model still achieved relatively high AUC on this task when response status was removed altogether. To get a clear picture of how our model performs without correctness information, we repeat the ablation study after removing response status from the input and skipping the response status pre-training objective.

\begin{table}[]
    \centering
    \begin{tabular}{|l|c|}
        \hline
        Remove & Test AUC \\
        \hline
        None & $0.792 \pm 0.004$ \\
        Event Type Prediction & $0.746 \pm 0.015$ \\
        Time Prediction & $0.794 \pm 0.008$ \\
        All Pre-Training Objectives & $0.774 \pm 0.005$ \\
        Attention & $0.747 \pm 0.017$ \\
        Fine-Tuning & $0.786 \pm 0.006$ \\
        \hline
    \end{tabular}
    \caption{We repeat the ablation experiment, but remove response status to examine how the model interacts with purely behavioral data.}
    \label{tab:ablation-score-no-rc}
\end{table}

We see in Table~\ref{tab:ablation-score-no-rc} that after removing response status and using only behavioral data to predict performance, all components of our framework are effective. The predictive performance does not suffer when the time prediction pre-training objective is skipped but suffers significantly when event type prediction is skipped. Interestingly, we see that the model performs better training from scratch than it does with the standard pre-training workflow without event type prediction. A possible explanation is that when we do not predict event types during pre-training, the latent states are overfitted to represent time information, which is not highly predictive of this label. Additionally, without response status, the model is much more dependent on attention. A possible explanation is that with response status, the final correctness state is always available at the end of the sequence, but without response status, relevant behavioral details may be lost throughout the sequence. Finally, we can see that the model benefits from fine-tuning, possibly because the latent states are not so directly indicative of student performance as they were when pre-trained on response status.

\subsection{Qualitative Analysis}

We now examine the interpretability of the latent behavioral vectors that our methodology produces. We will first examine question-level latent vectors extracted from the behavior-enhanced IRT model. We will then examine student-level latent vectors extracted from a version of the student-level model, which was modified to capture task-switching behavior. For the following figures, we use t-SNE \cite{t-sne} to visualize these vectors in a 2D plane, and investigate characteristic behavioral patterns in the visible clusters.

\paragraph{Question-level Vectors}

Using our model trained on the behavior-enhanced IRT task, we extract the representative vectors for student behavior across different questions. We color the point for each vector $\mathbf{b}^i_j$ based on the value of the associated behavior scalar $B_{ij}=\phi_p(\mathbf{b}^i_j)$; positive values increase the predicted probability of a correct response on top of the student ability and question difficulty parameters while negative values decrease it. We put positive and negative behavior scalars into separate groups, and then distinguish between points above or below the median in each group. We draw a vector with \checkmark  if the student got the question right and $\times$ if they got it wrong. Additionally, we randomly select a small subset of the vectors to show to avoid a cluttered figure. We circle a few clusters and examine trends of high-level features within each one, such as how many visits were made to the question or how much time was spent on it. We also examine the raw process data of representative points in clusters to identify low-level trends.

\begin{figure}
    \centering
    \includegraphics[width=3.3in]{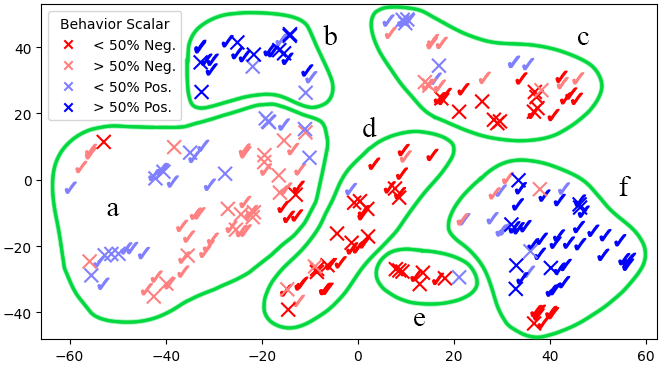}
    \caption{We show the question-level vectors extracted from the behavior-enhanced IRT model for a multiple choice question, marked by if the student got the question right or wrong, and colored by the behavior scalar assigned by the model. The vectors yielded distinct clusters, each representing separate patterns of problem-solving behaviors.}
    \label{fig:irt_qid_2}
\end{figure}

In Figure~\ref{fig:irt_qid_2}, we visualize the behavior vectors for an early multiple choice question in block A. We selected it for demonstration since its vectors displayed good clustering patterns. Within it, we identify 6 distinct clusters:
\begin{itemize}
    \item a - Few interactions: a majority of these students simply clicked an answer after looking at the question for less than a minute. A small number of them changed their answer, or engaged in some other interactions such as opening the calculator, but all are characterized by having relatively short sequences.
    \item b - Using the calculator and keeping it open: these students all used the calculator before clicking on a single answer and keeping the calculator open as they moved on to the next question; they also generally did not revisit the question. Our model assigned all these students a positive behavioral scalar, with most getting the question right, indicating this was a generally beneficial behavior type.
    \item c - Using the calculator and closing it: similar to the group to the left, all students used the calculator, and generally did not interact with items other than the calculator and the answers; the difference is that this group explicitly closed the calculator, sometimes opening and closing it multiple times within a visit. Some also visited the question multiple times. Compared to the group that kept the calculator open, the model assigned this group worse behavior scalars overall.
    \item d - Many visits: these students generally visited the question two or more times. A variety of event types were present in these sequences, including eliminating choices, drawing, highlighting, using the calculator, and changing answers. Each visit was also generally short in the number of events. The model almost universally assigned a negative value to this group.
    \item e - Text to speech: these students all used text to speech, and while some of them also drew and used the calculator, others responded without other interactions. The negative prediction bias for this group indicates the model may inadvertently demonstrate bias against students with disabilities, who are more likely to use text to speech. Investigating potential bias in the model, as well as mitigating its effect, is an important area of future work.
    \item f - High effort: compared to other groups, these students generally spent more time on the question and made more clicks; they also often made several visits to the question. The sequences were generally longer, and most of the events were associated with drawing and using the calculator, indicating the students spent significant effort finding a solution. The model generally assigned this group a positive value.
\end{itemize}

These clusters appear to represent distinct behavior types, with each one representing a unique combination of granular features, such as the use of specific helper tools, and high-level features, such as sequence length or number of visits. Different behavior types across clusters are generally quite distinguishable with limited overlap in some cases.

\paragraph{Student-level Vectors}

We now investigate student-level representations, which combine all question visits of a student into a single latent vector. Our goal is to explore whether these representations can capture task-switching behavior across questions in a time-limited NAEP assessment, which may capture different high-level task management strategies among students. This aspect of student behavior is not captured in CKT. We train a student-level model on the score label task with the following modifications to our original setup detailed above:
\begin{itemize}
    \item We provide the process model with sequences of events in a single visit to a question, rather than across all visits to a question.
    \item We add a new pre-training objective to predict the question ID of each event to encode task-switching information in the latent representations. This objective leads to better student-level representations.
    \item We replace the fully-connected neural network with a gated recurrent unit (GRU), an enhanced type of RNN, since the visits are sequential in nature. We use the final hidden state of the GRU as the input to a linear projection to predict the label. We also cluster this latent state as the student-level representation for the visualization.
\end{itemize}

\begin{figure}
    \centering
    \includegraphics[width=3in]{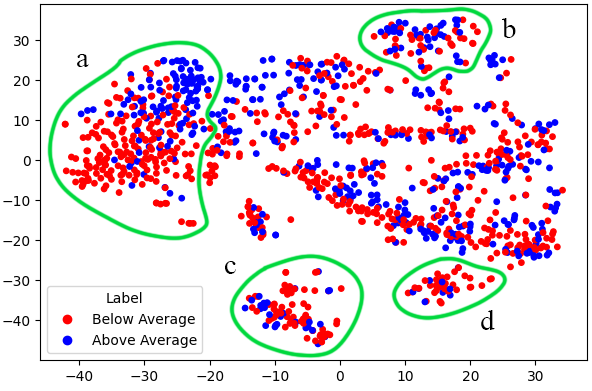}
    \caption{We show student-level vectors extracted from the final hidden state of a GRU that processes all sequential visit-level vectors from a student's process data in block A, colored by if the student performed above or below average in block B. Several regions represent distinct behavioral patterns over the course of block A, while others do not convey any clear behavioral distinctions.}
    \label{fig:student_level}
\end{figure}

In Figure~\ref{fig:student_level}, we visualize the student-level vectors, colored according to the score label. We identify 4 distinct clusters, while the rest of the vectors have no obvious pattern:
\begin{itemize}
    \item a - Rapid testing: most students in this group finished all questions in block A very quickly, often with 10-15 minutes remaining. This strategy is employed by both students with good and poor performance in block B; some students got most questions correct and are highly confident, while others rush through things without properly thinking about each question. 
    \item b - Checked their work: students in this group made multiple visits to most questions. In some cases they would make changes on the second visit, but in others they would leave their response unchanged. Most students in this group completed all questions.
    \item c and d - Ran out of time: both of these clusters represent students that ran out of time while completing block A, typically completing no more than half of the questions in the block. While most of these students perform poorly on block B as expected, several perform above average.
\end{itemize}

While the clusters for the student-level vectors are not as clearly distinguishable as those for the question-level vectors, we can still identify some clusters that correspond to clear task-switching behavior in online assessments. We note that there is a large group of students that have been not assigned a cluster; we either cannot identify any clear separation between these students and other clusters or cannot find particular patterns of their behavior. We think that adding pre-training objectives that are particularly designed to capture task-switching behavior may result in more distinguishable clusters in student-level vectors. We leave this investigation for future work.

\section{Conclusions}
In this paper, we developed a BERT-style framework for pre-training and transfer learning on educational process data. We applied our framework to several downstream learning outcome prediction tasks on NAEP assessment process data used in the NAEP 2019 data mining competition. Through quantitative and qualitative experiments, we demonstrated the effectiveness of our framework. We now discuss several important observations we made while developing the framework and lay out potential areas for future research.

First, while transfer learning on process data has been attempted before, we found that a BERT-style pre-training approach combined with an attention-based transfer function yields superior or similar downstream performance across tasks compared to existing methods. This observation implies that the representations produced by the BERT-like pre-training are robust and capable of benefiting different downstream tasks under a variety of conditions, and that attention is an effective technique for accessing these representations throughout the sequence. However, our process model currently only captures the context of single questions at a time, so a relevant area of future work would be expanding the framework to capture student sequences as a whole; this would enable the latent states to capture task-switching and different behavioral patterns throughout a student's process data, theoretically resulting higher performance on downstream tasks as well as more meaningful clusters of student representations. Advanced techniques may be needed to implement this effectively. One approach may be using a Transformer encoder model, which may be promising given the success of this architecture in NLP, but would require a much larger dataset for pre-training.

Second, we observed that clustering question-level and student-level representations revealed groups of different behavioral approaches to problem-solving and test-taking. This could be useful for identifying when students are utilizing, or failing to utilize, effective techniques. It may also provide pedagogical insight on what kinds of techniques are helpful across a variety of question types and student populations. However, the student-level representations resulted in clusters that were less distinguishable than the question-level representations, leaving room for future work to create more meaningful representations at this scale. Additionally, future work should investigate and seek to mitigate bias in the model against groups who may exhibit distinct behavioral patterns, such as students with physical or learning disabilities.

Finally, we note that while our method was successful on the NAEP dataset, we designed the methodology to handle any source of educational process data. Future work should implement our methodology in a different environment, such as video clickstreams, and compare its effectiveness to the setting explored in this paper. If successful, this will provide evidence that our methodology can capture meaningful aspects of student behavior in general, and may be expanded to improve performance on additional learning outcome tasks such as knowledge tracing.

\section{Acknowledgement}
We would like to thank the support of the National Science Foundation under grant IIS-1917713 and the UMass Interdisciplinary Faculty Research Award.

\balancecolumns

\newpage
\bibliographystyle{abbrv}
\bibliography{references}

\end{document}